\documentclass{article}
\usepackage[latin9]{inputenc}
\usepackage{amsmath}
\usepackage{amssymb}
\usepackage{stmaryrd}

\makeatletter

\providecommand{\tabularnewline}{\\}

\usepackage{spconf}
\usepackage{url}
\usepackage{graphicx}

\title{Indian Buffet Process Deep Generative Models for Semi-Supervised Classification}
\name{Sotirios P. Chatzis}
\address{Department of Electrical Eng., Computer Eng., and Informatics\\
Cyprus University of Technology\\ Limassol 3036, Cyprus}
\makeatother

\begin{document}
\maketitle 
\begin{abstract}
Deep generative models (DGMs) have brought about a major breakthrough,
as well as renewed interest, in generative latent variable models.
However, DGMs do not allow for performing data-driven inference of
the number of latent features needed to represent the observed data.
Traditional linear formulations address this issue by resorting to
tools from the field of nonparametric statistics. Indeed, linear latent
variable models imposed an Indian Buffet Process (IBP) prior have
been extensively studied by the machine learning community; inference
for such models can been performed either via exact sampling or via
approximate variational techniques. Based on this inspiration, in
this paper we examine whether similar ideas from the field of Bayesian
nonparametrics can be utilized in the context of modern DGMs in order
to address the latent variable dimensionality inference problem. To
this end, we propose a novel DGM formulation, based on the imposition
of an IBP prior. We devise an efficient Black-Box Variational inference
algorithm for our model, and exhibit its efficacy in a number of semi-supervised
classification experiments. In all cases, we use popular benchmark
datasets, and compare to state-of-the-art DGMs.
\end{abstract}

\section{Introduction}

Linear latent variable (LLV) models, including, among others, factor
analysis (FA) and probabilistic principal component analysis (PPCA),
have a long tradition in the field of generative modeling of high-dimensional
observations with underlying latent structure. One of the difficulties
related with the utilization of LLV models concerns the determination
of the most appropriate number of latent variables (latent vector
dimensionality) for representing a given dataset, without resorting
to cross-validation. To this end, several researchers have considered
utilization of concepts from the field of Bayesian nonparametrics. 

Nonparametric Bayesian models postulate a (theoretically) infinite-dimensional
latent variable space. Appropriate priors are imposed over the postulated
(infinite-dimensional) latent variables, that allow for deriving effective,
data-driven posterior distributions over the latent dimension generation
process. Specifically, nonparametric formulations of LLV models are
most often obtained by imposition of an Indian Buffet Process (IBP)
prior over the model latent variables \cite{cikm}. The IBP prior
\cite{ibp} is a nonparametric prior for latent feature models where
observations are influenced by a combination of hidden features. It
offers a principled prior in diverse contexts where the number of
latent features is unknown. Its rationale consists in eventually utilizing
only a finite set of ``effective'' latent variables to represent
the observed data points. This set is determined in a heuristics-free,
data-driven way, as a part of the resulting inference algorithm \cite{cikm}.

Despite these advances, the linear assumptions of LLV models cannot
be considered realistic in most real-world data modeling scenarios.
As such, in the last couple of years, immense research interest has
concentrated on the development of nonlinear latent variable models,
where the inferred latent variable posteriors are parameterized via
deep neural networks. This novel class of latent variable models is
commonly referred to as deep generative models (DGMs) \cite{aevb,gershman}. 

Inspired from these advances, in this paper we address the problem
of automatic data-driven inference of the latent variable dimensionality
in DGMs. Specifically, we examine whether a nonparametric Bayesian
formulation of DGMs, based on the utilization of the IBP prior, would
offer an attractive solution to this problem. To this end, we devise
a novel nonparametric hierarchical graphical formulation of DGMs,
whereby the observed data are described via a factorized latent variable
construction, driven by some latent indicators of data point allocation
which are imposed an IBP prior. We derive an efficient inference algorithm
for our model by resorting to Black-Box Variational Inference (BBVI)
\cite{bbvi,bbvi2}. 

The remainder of this paper is organized as follows: In Section 2,
we briefly outline the methodological background of our approach.
In Section 3, we introduce our approach and derive its inference algorithms.
In Section 4, we perform a thorough experimental evaluation, using
benchmark data. Finally, in the concluding Section, we briefly summarize
our results.

\section{Theoretical Background}

\subsection{DGMs}

In their \emph{basic formulation,} DGMs assume that the observed random
variables $\boldsymbol{x}$ are generated by some random process,
involving an unobserved \emph{continuous} \emph{random} \emph{vector}
$\boldsymbol{z}$, with some prior distribution $p(\boldsymbol{z})$.
The observed variables $\boldsymbol{x}$ are considered i.i.d. given
the corresponding latent variables $\boldsymbol{z}$, with conditional
likelihood function $p(\boldsymbol{x}|\boldsymbol{z};\boldsymbol{\theta})$.
This way, the model's log-marginal likelihood can be lower-bounded
as (evidence lower bound, ELBO): 
\begin{equation}
\mathrm{log}\,p(\boldsymbol{x})\geq\mathcal{L}(\boldsymbol{\phi})=\mathbb{E}_{q(\boldsymbol{z};\boldsymbol{\phi})}[\mathrm{log}\,p(\boldsymbol{x},\boldsymbol{z})-\mathrm{log}\,q(\boldsymbol{z};\boldsymbol{\phi})]
\end{equation}
where $\mathbb{E}_{q(\boldsymbol{z};\boldsymbol{\phi})}[\cdot]$ is
the expectation of a function w.r.t. the random variable $\boldsymbol{z}$,
drawn from $q(\boldsymbol{z};\boldsymbol{\phi})$, and $q(\boldsymbol{z};\boldsymbol{\phi})$
is the approximate (variational) posterior over the latent variable
$\boldsymbol{z}$, that is inferred from the data.

DGMs assume that the likelihood function of the model, $\mathrm{log}\,p(\boldsymbol{x}|\boldsymbol{z};\boldsymbol{\theta})$,
as well as the inferred approximate (variational) latent variable
posterior, $q(\boldsymbol{z};\boldsymbol{\phi})$, are parameterized
via deep neural networks. For computational efficiency, $q(\boldsymbol{z};\boldsymbol{\phi})$
is typically taken as a diagonal Gaussian: 
\begin{equation}
q(\boldsymbol{z};\boldsymbol{\phi})=\mathcal{N}(\boldsymbol{z}|\boldsymbol{\mu}(\boldsymbol{x};\boldsymbol{\phi}),\mathrm{diag}\;\boldsymbol{\sigma}^{2}(\boldsymbol{x};\boldsymbol{\phi}))
\end{equation}
where the $\boldsymbol{\mu}(\boldsymbol{x};\boldsymbol{\phi})$ and
$\boldsymbol{\sigma}^{2}(\boldsymbol{x};\boldsymbol{\phi})$ are outputs
of deep neural networks, and $\mathrm{diag}\,\boldsymbol{\chi}$ is
a diagonal matrix with $\boldsymbol{\chi}$ on its main diagonal.
Under these assumptions, variational (approximate) inference is performed
by drawing Monte Carlo samples from $q(\boldsymbol{z};\boldsymbol{\phi})$,
which are further reparameterized as deterministic functions of the
posterior mean $\boldsymbol{\mu}(\boldsymbol{x};\boldsymbol{\phi})$,
variance $\boldsymbol{\sigma}^{2}(\boldsymbol{x};\boldsymbol{\phi})$,
and some white random noise variable $\boldsymbol{\epsilon}$ \cite{aevb}:
\begin{equation}
\boldsymbol{z}=\boldsymbol{\mu}(\boldsymbol{x};\boldsymbol{\phi})+\boldsymbol{\sigma}(\boldsymbol{x};\boldsymbol{\phi})\varodot\boldsymbol{\epsilon},\;\;\mathrm{with}\;\boldsymbol{\epsilon}\sim\mathcal{N}(\boldsymbol{0},\boldsymbol{I})
\end{equation}
where $\varodot$ is the elementwise product between vectors. Specifically,
these samples are used to approximate the intractable posterior expectations
in (1), in a way that results in low-variance estimators, $\boldsymbol{\phi}$,
of the sought posterior, $q(\boldsymbol{z};\boldsymbol{\phi})$ \cite{aevb}.

\subsection{Nonparametric Modeling Using the IBP Prior}

The IBP is a prior on infinite binary matrices that allows us to simultaneously
infer which features influence a set of observations and how many
features there are. The form of the prior ensures that only a finite
number of features will be present in any finite set of observations,
but more features may appear as more observations are received. Let
us consider a set of $N$ objects that may be assigned to a total
of $K\rightarrow\infty$ features. Let $\boldsymbol{Z}=[z_{ik}]_{i,k=1}^{N,K}$
be a $N\times K$ matrix of assignment variables, with $z_{ik}=1$
if the $i$th object is assigned to the $k$th feature (multiple $z_{ik}$'s
may be equal to 1 for a given object $i$), $z_{ik}=0$ otherwise.
Then, a formulation of the IBP that renders $p(\boldsymbol{Z})$ amenable
to variational inference consists in the following hierarchical representation
\cite{vbibp}:

\begin{equation}
z_{ik}\sim\mathrm{Bernoulli}(\pi_{k})\,\forall i
\end{equation}
\begin{equation}
\pi_{k}=\prod_{j=1}^{k}v_{j},\;v_{k}\sim\mathrm{Beta}(\alpha,1)\;\forall k
\end{equation}

\subsection{BBVI}

BBVI is an effective means of performing variational inference for
DGM variants that entail \emph{discrete random variables}. Let us
consider a probabilistic model $p(\boldsymbol{x},\boldsymbol{z})$
and a sought variational family $q(\boldsymbol{z};\boldsymbol{\phi})$.
BBVI optimizes the ELBO (1) by relying on the ``log-derivative trick\textquotedblright{}
\cite{reinforce} to obtain Monte Carlo estimates of the gradient
that reads
\begin{equation}
\nabla_{\boldsymbol{\phi}}\mathcal{L}(\boldsymbol{\phi})=\mathbb{E}_{q(\boldsymbol{z};\boldsymbol{\phi})}[f(\boldsymbol{z})]
\end{equation}
where 
\begin{equation}
f(\boldsymbol{z})=\nabla_{\boldsymbol{\phi}}\mathrm{log}\,q(\boldsymbol{z};\boldsymbol{\phi})\left[\mathrm{log}\,p(\boldsymbol{x},\boldsymbol{z})-\mathrm{log}\,q(\boldsymbol{z};\boldsymbol{\phi})\right]
\end{equation}
Then, to reduce the variance of the estimator, one common strategy
in BBVI consists in the use of \emph{control variates. }A control
variate is a random variable that is included in the estimator, preserving
its expectation but reducing its variance. The most usual choice for
control variates, which we adopt in this work, is the so-called weighted
score function: Under this selection, the ELBO gradient becomes 
\begin{equation}
\nabla_{\boldsymbol{\phi}}\mathcal{L}(\boldsymbol{\phi})=\sum_{n=1}^{N}\mathbb{E}_{q(\boldsymbol{z};\boldsymbol{\phi})}[f_{n}(\boldsymbol{z})-a_{n}h_{n}(\boldsymbol{z})]
\end{equation}
where $f_{n}(\cdot)$ and $h_{n}(\cdot)$ are the $n$th component
of $f(\cdot)$ and $h(\cdot)$, respectively, we denote 
\begin{equation}
h_{n}(\boldsymbol{z})=\nabla_{\boldsymbol{\phi}}\mathrm{log}\,q(\boldsymbol{z}_{n};\boldsymbol{\phi})
\end{equation}
and the constants $a_{n}$ are given by \cite{bbvi} 
\begin{equation}
a_{n}=\frac{\mathrm{Cov}\left(f_{n}(\boldsymbol{z}),h_{n}(\boldsymbol{z})\right)}{\mathrm{Var}\left(h_{n}(\boldsymbol{z})\right)}
\end{equation}
On this basis, derivation of the sought variational posteriors is
performed by utilizing the gradient expression (8) in the context
of off-the-shelf stochastic gradient optimizers. Specifically, in
this work we utilize AdaM \cite{adam}.

\section{Proposed Approach}

Let us consider the dataset $X=\{\boldsymbol{x}_{i}\}_{i=1}^{N}$.
The proposed IBP-DGM model assumes a conditional likelihood $p(\boldsymbol{x}_{i}|\boldsymbol{z}_{i};\boldsymbol{\theta})$,
parameterized by deep neural networks, and selected similar to the
case of conventional DGMs; for instance, in case of real observations,
$\boldsymbol{x}_{i}\in\mathbb{R}^{D}$, a diagonal Gaussian likelihood
is selected; in cases of binary observations, $\boldsymbol{x}_{i}\in\{0,1\}^{D}$,
we opt for a Bernoulli likelihood. Further, we introduce the following
hierarchical prior formulation for the latent variables $\boldsymbol{z}_{i}$:
\begin{equation}
\boldsymbol{z}_{i}=\tilde{\boldsymbol{z}}_{i}\cdot\hat{\boldsymbol{z}}_{i}
\end{equation}
\begin{equation}
p(\tilde{\boldsymbol{z}}_{i})=\mathcal{N}(\tilde{\boldsymbol{z}}_{i}|\boldsymbol{0},\boldsymbol{I})
\end{equation}
\begin{equation}
p(\hat{\boldsymbol{z}}_{i})=\prod_{k=1}^{K\rightarrow\infty}\mathrm{Bernoulli}(\hat{z}_{ik}|\pi_{k})
\end{equation}
\begin{equation}
\pi_{k}\triangleq\prod_{j=1}^{k}v_{j},\;k\in\{1,\dots,\infty\}
\end{equation}
\begin{equation}
p(v_{k})=\mathrm{Beta}(v_{k}|\alpha,1),\;k\in\{1,\dots,\infty\}
\end{equation}
The introduction of the binary latent variables $\hat{\boldsymbol{z}}_{i}$
in Eq. (11) essentially allows for the model to infer which latent
features $\tilde{z}_{ik},\;k\in\{1,\dots,K\rightarrow\infty\}$, are
active for each one of the observed data $\boldsymbol{x}_{i}$. This
way, if a latent feature, say the $k$th, yields drawn samples of
the indicators $\hat{z}_{ik}$ that are equal to zero for every observation,
$\boldsymbol{x}_{i}$, it will be effectively ignored by the model.

Under the infinite dimensional setting prescribed in Eqs. (11)-(15),
Bayesian inference is not feasible. For this reason, we employ a common
strategy in the literature of Bayesian nonparametrics, formulated
on the basis of a truncated, implicitly finite, representation of
the IBP \cite{vbibp}. That is, we fix a value $K\ll\infty$, letting
the posterior over the $v_{k}$ have the property $q(v_{K}=0)=1$.
In other words, we set the $\pi_{k}$ equal to zero for $k>K$ $\forall i$.
We then postulate: 
\begin{equation}
q(\tilde{\boldsymbol{z}}_{i};\boldsymbol{\phi})=\mathcal{N}(\tilde{\boldsymbol{z}}_{i}|\boldsymbol{\mu}(\boldsymbol{x}_{i};\boldsymbol{\phi}),\mathrm{diag}\;\boldsymbol{\sigma}^{2}(\boldsymbol{x}_{i};\boldsymbol{\phi}))
\end{equation}
\begin{equation}
q(\hat{\boldsymbol{z}}_{i};\boldsymbol{\phi})=\prod_{k=1}^{K}\mathrm{Bernoulli}(\hat{z}_{ik}|\hat{\pi}_{k}(\boldsymbol{x}_{i};\boldsymbol{\phi}))
\end{equation}
\begin{equation}
q(v_{k};\boldsymbol{\phi})=\mathrm{Beta}(v_{k}|a_{k}(\boldsymbol{x}_{i};\boldsymbol{\phi}),b_{k}(\boldsymbol{x}_{i};\boldsymbol{\phi})),\;k\in\{1,\dots,K\}
\end{equation}
Here, the $\boldsymbol{\mu}(\boldsymbol{x}_{i};\boldsymbol{\phi})$,
$\boldsymbol{\sigma}^{2}(\boldsymbol{x}_{i};\boldsymbol{\phi})$,
$\hat{\pi}_{k}(\boldsymbol{x}_{i};\boldsymbol{\phi})$, $a_{k}(\boldsymbol{x}_{i};\boldsymbol{\phi})$,
and $b_{k}(\boldsymbol{x}_{i};\boldsymbol{\phi})$ are parameterized
by deep neural networks. Finally, we impose a simple spherical prior
over the likelihood parameters $\boldsymbol{\theta}$: 
\begin{equation}
p(\boldsymbol{\theta})=\mathcal{N}(\boldsymbol{\theta}|\boldsymbol{0},\sigma_{\theta}^{2}\boldsymbol{I})
\end{equation}
In addition, to facilitate computational efficiency, we consider that
the sought variational posterior $q(\boldsymbol{\theta})$ collapses
to a single point, $\hat{\boldsymbol{\theta}}$, that essentially
constitutes a point-estimate; i.e., $q(\boldsymbol{\theta})=\delta_{\hat{\boldsymbol{\theta}}}(\boldsymbol{\theta})$,
where $\delta_{\hat{\boldsymbol{\theta}}}(\boldsymbol{\theta})$ is
a distribution over $\boldsymbol{\theta}$ with all its mass concentrated
on $\hat{\boldsymbol{\theta}}$.

This concludes the formulation of the IBP-DGM model. Even though IBP-DGM
is a generative model, we can use it to perform semi-supervised learning.
To this end, we only need to modify the likelihood function so as
to take into account (possible) label information. Specifically, we
postulate a different class-conditional likelihood function for each
class label, $y$, of the form $p(\boldsymbol{x}|\tilde{\boldsymbol{z}}\cdot\hat{\boldsymbol{z}},y;\boldsymbol{\theta})$;
this is employed for all the labeled training data points belonging
to the corresponding class. On the other hand, we continue to use
the likelihood function $p(\boldsymbol{x}|\tilde{\boldsymbol{z}}\cdot\hat{\boldsymbol{z}};\boldsymbol{\theta})$
for the available unlabeled data points. Finally, we also need to
introduce a prior $p(y)$ over the labels $y$ of the observed data,
as well as a corresponding variational posterior $q(y;\boldsymbol{\phi})$.
We have 
\begin{equation}
p(y=c)=\frac{1}{C},\;\forall c
\end{equation}
and 
\begin{equation}
q(y;\boldsymbol{\phi})=\mathrm{Cat}(y|\varpi(\boldsymbol{x}))
\end{equation}
where $\varpi(\boldsymbol{x})$ is parameterized via a deep network,
and $C$ is the total number of classes.

Then, variational inference is performed by resorting to BBVI, as
described in the previous Section. The ELBO expression of the model
reads:
\begin{equation}
\begin{aligned}\mathcal{L}(\boldsymbol{\phi};\boldsymbol{\theta})= & \mathbb{E}_{q(\tilde{\boldsymbol{z}};\boldsymbol{\phi})}[p(\tilde{\boldsymbol{z}})-\mathrm{log}\,q(\tilde{\boldsymbol{z}};\boldsymbol{\phi})]\\
+ & \mathbb{E}_{q(\hat{\boldsymbol{z}};\boldsymbol{\phi})}[p(\hat{\boldsymbol{z}})-\mathrm{log}\,q(\hat{\boldsymbol{z}};\boldsymbol{\phi})]\\
+ & \mathbb{E}_{q(y;\boldsymbol{\phi})}[p(y)-\mathrm{log}\,q(y;\boldsymbol{\phi})]\\
+ & \sum_{\boldsymbol{x}_{i}:y_{i}=y}\mathbb{E}_{q(\tilde{\boldsymbol{z}},\hat{\boldsymbol{z}};\boldsymbol{\phi})}[p(\boldsymbol{x}_{i}|\tilde{\boldsymbol{z}}_{i}\cdot\hat{\boldsymbol{z}}_{i},y;\boldsymbol{\theta})]\\
+ & \sum_{\boldsymbol{x}_{i}:y_{i}=\emptyset}\mathbb{E}_{q(\tilde{\boldsymbol{z}},\hat{\boldsymbol{z}};\boldsymbol{\phi})}[p(\boldsymbol{x}_{i}|\tilde{\boldsymbol{z}}_{i}\cdot\hat{\boldsymbol{z}}_{i};\boldsymbol{\theta})]
\end{aligned}
\end{equation}

\section{Experiments}

To exhibit the efficacy of our approach, we perform evaluation using
the MNIST, Rotated MNIST+Background Images, MNIST+Background Images,
MNIST+Random Background, Rotated MNIST, and (Small-)NORB benchmarks.\footnote{Before each epoch, the normalized MNIST images are binarized by sampling
Bernoulli distributions, similar to \cite{mnistb}. We normalize all
NORB images following the procedure suggested in \cite{vat}; we add
uniform noise between 0 and 1 to each pixel value, to allow for effectively
modeling them by means of Gaussian conditional likelihoods, $p(\boldsymbol{x}|\tilde{\boldsymbol{z}}\cdot\hat{\boldsymbol{z}},y;\boldsymbol{\theta})$
and $p(\boldsymbol{x}|\tilde{\boldsymbol{z}}\cdot\hat{\boldsymbol{z}};\boldsymbol{\theta})$. } We perform evaluations under an experimental setup where $1\%$ of
the available training data is presented to the trained models as
\emph{labeled} training examples (randomly selected, in equal proportions
from each class), while the rest is used as \emph{unlabeled} training
examples. To provide some comparative results, apart from our method
we also evaluate the M2 approach proposed in \cite{aevb2}, which
constitutes the parametric equivalent of IBP-DGM in the context of
semi-supervised learning. We consider two alternative architectures
of the deep networks parameterizing the postulated likelihood and
posterior distributions of IBP-DGM. The first alternative comprises
simple Dense Layer (DL) architectures. The second one is based on
the Memory Network (MN) architecture recently proposed in \cite{mn}.
This employs an external hierarchical memory to capture variant information
at different abstraction levels trained in an unsupervised manner. 

In all our experiments, for simplicity and computational efficiency,
we use architectures comprising only one hidden layer (DL or MN),
with 500 (deterministic) units. We use ReLU nonlinearities for all
the postulated (deterministic) hidden units \cite{relu}. Initialization
of the network parameters is performed by adopting a Glorot-style
uniform initialization scheme \cite{glorot}. The used MN layers comprise
100 memory slots; that is the number of rows of matrix $\boldsymbol{A}$,
or, conversely, the number of columns of the memory matrix $\boldsymbol{M}$
\cite{mn}. In all cases, the maximum size of the postulated latent
vectors $\boldsymbol{z}$ (truncation threshold $K$ of the variational
posterior) is set to 50.\footnote{In each case, the prior variance $\sigma_{\theta}^{2}$ of the model
parameters $\boldsymbol{\theta}$ is heuristically selected among
the alternative values $\{10^{-3},10^{-2},10^{-1}\}$, with the aim
of maximizing out-of-sample predictive performance. To execute AdaM,
we use a learning rate of $3\times10^{-4}$, and an exponential decay
rate for the first and second moment at 0.9 and 0.999, respectively. } Our source codes have been developed in Python, and make use of the
Tensorflow library \cite{tensorflow2015-whitepaper}.

In Tables 1 and 2, we provide the obtained performance results (error
rates \%) of the evaluated methods under the two considered experimental
scenarios. These figures are average performance results over 50 repetitions
of our experiments, with different random training data splits into
labeled and unlabeled subsets each time. As we observe, our approach
yields a clear improvement over the competition in all cases. To examine
the statistical significance of the observed performance differences,
we run the Student's-$t$ statistical significance test on the pairs
of performances of our method and M2. The test rejected the null hypothesis,
with $p$-values below $10^{-8}$, in all cases.

Another interesting observation is that the obtained improvement of
IBP-DGM over M2 is more profound in the case of the DL parameterization.
We suspect this result is due to the fact that the MN parameterization
introduces an attention mechanism which essentially puts more or less
emphasis on some latent characteristics of the data. This might turn
out to be more beneficial for some parametric model than for a nonparametric
one, which already includes a (different sort of) mechanism for latent
feature retention or omission.

Note also that IBP-DGM requires similar computational time to generate
one prediction compared to the competition. Turning to the training
algorithm of our approach, we can report the following quite interesting
finding: When using the DL parameterization, IBP-DGM requires approximately
4 times more algorithm epochs to converge compared to one M2 network;
this is the case for all the considered benchmarks. On the other hand,
when using the MN parameterization, both approaches require similar
numbers of epochs to converge; this is approximately 4 times more
epochs compared to one M2 network with DL parameterization. Our interpretation
of this finding is that the introduction of a mechanism that puts
less or more emphasis on some latent features requires that model
training proceeds more slowly.

\begin{table}
\caption{Semi-supervised test error (\%) using the considered DL parameterization.}

\centering{}{\small{}}%
\begin{tabular}{|c|c|c|}
\hline 
{\small{}Method } & {\small{}M2 } & {\small{}IBP-DGM}\tabularnewline
\hline 
\hline 
{\small{}MNIST } & {\small{}8.10 } & {\small{}7.85}\tabularnewline
\hline 
{\small{}Rotated MNIST } & {\small{}38.80 } & {\small{}32.82}\tabularnewline
\hline 
{\small{}MNIST+Background Images } & {\small{}16.16 } & {\small{}8.99}\tabularnewline
\hline 
{\small{}MNIST+Random Background } & {\small{}12.34 } & {\small{}7.78}\tabularnewline
\hline 
{\small{}Rotated MNIST+Background Images } & {\small{}12.69 } & {\small{}8.03}\tabularnewline
\hline 
{\small{}NORB } & {\small{}18.02 } & {\small{}15.14}\tabularnewline
\hline 
\end{tabular}{\small\par}
\end{table}

\begin{table}
\caption{Semi-supervised test error (\%) using the considered MN parameterization.}

\centering{}{\small{}}%
\begin{tabular}{|c|c|c|}
\hline 
{\small{}Method } & {\small{}M2 } & {\small{}IBP-DGM}\tabularnewline
\hline 
\hline 
{\small{}MNIST } & {\small{}8.04 } & {\small{}7.45}\tabularnewline
\hline 
{\small{}Rotated MNIST } & {\small{}37.29 } & {\small{}32.80}\tabularnewline
\hline 
{\small{}MNIST+Background Images } & {\small{}9.08 } & {\small{}7.94}\tabularnewline
\hline 
{\small{}MNIST+Random Background } & {\small{}7.87 } & {\small{}6.85}\tabularnewline
\hline 
{\small{}Rotated MNIST+Background Images } & {\small{}8.42 } & {\small{}7.95}\tabularnewline
\hline 
{\small{}NORB } & {\small{}15.57 } & {\small{}14.88}\tabularnewline
\hline 
\end{tabular}{\small\par}
\end{table}

Finally, it is interesting to examine the values of the posteriors
over the latent indicators, $q(\hat{z}_{\cdot k};\boldsymbol{\phi})$,
obtained in each one of the previously considered experimental scenarios.
As we have observed, our model tends to yield high enough posterior
values only for the first 10-12 latent components. In most cases,
the posterior values, $q(\hat{z}_{\cdot k};\boldsymbol{\phi})$, of
the active components tend to yield higher mean values, and most importantly,
higher standard deviations, in the case of the DL parameterization.
In our view, this outcome vouches for our previous claims that the
attention mechanisms of the MN network are actually complementary
to the nonparametric feature omission/retention mechanisms of the
IBP prior: When both mechanisms are used, they tend to reinforce each
other. This results in a lower standard deviation for the $q(\hat{z}_{\cdot k};\boldsymbol{\phi})$
values of the active components across the training data points.

\section{Conclusions}

In this paper, we addressed the problem of performing inference over
the latent variable dimensionality of DGMs. To this end, we devised
a nonparametric formulation of DGMs, obtained by utilizing the IBP
prior. We performed inference for the so-derived IBP-DGM model by
resorting to the BBVI inference scheme. As we showed, our approach
is quite effective in terms of inferring the latent variable dimensionality,
and yields competitive classification performance. Remarkably, the
observed modeling and predictive performance improvement did not come
at the cost of extra computational overheads. Our future work will
focus on extending DGMs so as to model heteroscedastic data, e.g.
\cite{gpmch,tnnls12}, as well as data with temporal dynamics of unknown
order, e.g. \cite{iCRF}.

\bibliographystyle{IEEEbib}
\bibliography{ibp-mem}

\begin{thebibliography}{10}

\bibitem{cikm}
Sotirios~P. Chatzis,
\newblock ``Nonparametric {Bayesian} multitask collaborative filtering,''
\newblock in {\em Proc. ACM CIKM}, 2013.

\bibitem{ibp}
T.~Griffiths and Z.~Ghahramani,
\newblock ``Infinite latent feature models and the {Indian} buffet process,''
\newblock Tech. {R}ep. TR 2005-001, Gatsby Computational Neuroscience Unit,
  2005.

\bibitem{aevb}
D.~Kingma and M.~Welling,
\newblock ``Auto-encoding variational {Bayes},''
\newblock in {\em Proc. ICLR}, 2014.

\bibitem{gershman}
S.~J. Gershman and N.~D. Goodman,
\newblock ``Amortized inference in probabilistic reasoning,''
\newblock in {\em Proc. Annual Conference of the Cognitive Science Society},
  2014.

\bibitem{bbvi}
Rajesh Ranganath, Sean Gerrish, and David~M. Blei,
\newblock ``Black box variational inference,''
\newblock in {\em Proc. AISTATS}, 2014.

\bibitem{bbvi2}
Francisco J.~R. Ruiz, Michalis~K. Titsias, and David~M. Blei,
\newblock ``Overdispersed black-box variational inference,''
\newblock in {\em Proc. UAI}, 2016.

\bibitem{vbibp}
Finale Doshi-Velez, Kurt Miller, Jurgen~Van Gael, and Yee~Whye Teh,
\newblock ``Variational inference for the {Indian Buffet Process},''
\newblock in {\em Proc. AISTATS}, 2009.

\bibitem{reinforce}
R.~J. Williams,
\newblock ``Simple statistical gradient-following algorithms for connectionist
  reinforcement learning,''
\newblock {\em Machine Learning}, vol. 8, no. 3-4, pp. 229--256, 1992.

\bibitem{adam}
Diederik Kingma and Jimmy Ba,
\newblock ``Adam: A method for stochastic optimization,''
\newblock in {\em Proc. ICLR}, 2015.

\bibitem{mnistb}
B.~Uria, I.~Murray, and H.~Larochelle,
\newblock ``A deep and tractable density estimator,''
\newblock in {\em Proc. ICML}, 2014.

\bibitem{vat}
T.~Miyato, S.~Maeda, M.~Koyama, K.~Nakae, and S.~Ishii,
\newblock ``Distributional smoothing with virtual adversarial training,''
\newblock in {\em Proc. ICLR}, 2016.

\bibitem{aevb2}
D.~P. Kingma, D.~J. Rezende, S.~Mohamed, and M.~Welling,
\newblock ``Semi-supervised learning with deep generative models,''
\newblock in {\em Proc. NIPS}, 2014.

\bibitem{mn}
Chongxuan Li, Jun Zhu, and Bo~Zhang,
\newblock ``Learning to generate with memory,''
\newblock in {\em Proc. ICML}, 2016.

\bibitem{relu}
Vinod Nair and Geoffrey~E. Hinton,
\newblock ``Rectified linear units improve restricted {Boltzmann} machines,''
\newblock in {\em Proc. ICML}, 2010.

\bibitem{glorot}
X.~Glorot and Y.~Bengio,
\newblock ``Understanding the difficulty of training deep feedforward neural
  networks,''
\newblock in {\em Proc. AISTATS}, 2010.

\bibitem{tensorflow2015-whitepaper}
Mart\'{\i}n Abadi et~al.,
\newblock ``{TensorFlow}: Large-scale machine learning on heterogeneous
  systems,'' 2015,
\newblock Software available from tensorflow.org.

\bibitem{gpmch}
Emmanouil~A. Platanios and Sotirios~P. Chatzis,
\newblock ``Gaussian process-mixture conditional heteroscedasticity,''
\newblock {\em IEEE Transactions on Pattern Analysis and Machine Intelligence},
  vol. 36, no. 5, pp. 888--900, 2014.

\bibitem{tnnls12}
Sotirios~P. Chatzis and Y.~Demiris,
\newblock ``Nonparametric mixtures of {Gaussian} processes with power-law
  behavior,''
\newblock {\em IEEE Transactions on Neural Networks and Learning Systems}, vol.
  23, pp. 1862--1871, Dec. 2012.

\bibitem{iCRF}
Sotirios~P. Chatzis and Yiannis Demiris,
\newblock ``The infinite-order conditional random field model for sequential
  data modeling,''
\newblock {\em IEEE Transactions on Pattern Analysis and Machine Intelligence},
  vol. 35, no. 6, pp. 1523--1534, 2013.

\end{thebibliography}

\end{document}